\PassOptionsToPackage{table,dvipsnames}{xcolor}
\documentclass[11pt,letterpaper]{article}

\usepackage{acl2023}

\definecolor{tfidfcolor}{HTML}{152EFF}    
\definecolor{poscolor}{HTML}{E67300}     
\definecolor{nercolor}{HTML}{6D904F}     
\definecolor{embedcolor}{HTML}{30A2DA}   
\definecolor{hybridcolor}{HTML}{BCBD22}  

\definecolor{lightrow}{gray}{0.90}       
\definecolor{highlightrow}{gray}{0.80}   

\usepackage{times}
\usepackage{latexsym}
\usepackage{amsmath}
\usepackage{amssymb}
\usepackage{graphicx}
\usepackage{booktabs}
\usepackage{enumitem}
\usepackage{tabularx}
\usepackage{float}
\usepackage{microtype}

\usepackage[T1]{fontenc}
\usepackage[utf8]{inputenc}
\usepackage{inconsolata}

\usepackage{listings}
\usepackage{tikz}
\usetikzlibrary{arrows.meta, positioning, shapes.geometric, patterns}
\tikzset{arrow/.style={-{Stealth}, thick}}

\title{LinguaSynth: Heterogeneous Linguistic Signals for News Classification}
\author{%
  Duo Zhang \quad Junyi Mo \\
  New York University \\
  \texttt{\{dz2349,jm9847\}@nyu.edu}
}
\date{}

\begin{document}
\maketitle

\begin{abstract}
Deep learning has significantly advanced NLP, but its reliance on large, black-box models introduces critical interpretability and computational efficiency concerns. This paper proposes LinguaSynth, a novel text classification framework that strategically integrates five complementary linguistic feature types—lexical, syntactic, entity-level, and both word-level and document-level semantics—within a transparent logistic regression model. Unlike transformer-based architectures, LinguaSynth maintains interpretability and computational efficiency, achieving an accuracy of 84.89\% on the 20 Newsgroups dataset, surpassing a robust TF-IDF baseline by 3.32\%. Through rigorous feature interaction analysis, we show that syntactic and entity-level signals provide essential disambiguation, effectively complementing distributional semantics. LinguaSynth sets a new benchmark for interpretable, resource-efficient NLP models, challenging the prevailing assumption that deep neural networks are necessary for high-performing text classification. 
\end{abstract}

\section{Introduction}
Deep learning has revolutionized NLP, but its success comes with trade-offs: high computational cost, limited interpretability, and increasing environmental impact. These challenges hinder adoption in regulated, resource-constrained, and sustainability-focused domains, raising a crucial question: Are large neural networks always necessary, or can simpler, transparent models suffice? 

We introduce LinguaSynth, a novel framework that demonstrates how strategic linguistic feature engineering, rather than algorithmic complexity, can drive strong classification performance. LinguaSynth integrates five complementary feature types:

\begin{enumerate}[itemsep=0pt, topsep=0pt, parsep=0pt, partopsep=0pt]
\item \textbf{Lexical features} (TF-IDF) - surface-level word importance
\item \textbf{Syntactic patterns} (part-of-speech histograms, bigrams) - grammatical structure. 
\item \textbf{Entity-level information} (named entity recognition) - named entity cues. 
\item \textbf{Word-level semantics} (GloVe) - local distributional meaning.
\item \textbf{Document-level semantics} (Doc2Vec) – global thematic content.
\end{enumerate}

Each feature type captures a distinct linguistic dimension, and their complementarity explains the significant performance gains observed. While components like TF-IDF or POS tagging are well-established, their strategic fusion within a transparent, balanced architecture is a novel contribution with broad applicability. 

On the 20 Newsgroups dataset, F1 scores improve in 19 out of 20 categories. Our experiments show that augmenting TF-IDF with distributional semantics (GloVe, Doc2Vec), syntax (POS), and entity signals (NER) leads to substantial gains. Although GloVe and Doc2Vec underperform in isolation, combining them with TF-IDF yields a +2.87\% lift in test accuracy, and further integrating POS and NER pushes performance to 84.89\%, surpassing the TF-IDF baseline by +3.32\%. 

Our analysis uncovers clear category-specific patterns that help explain why fusion-based models consistently outperform single-feature approaches. Real-world text categories exhibit unique linguistic signatures distributed across multiple dimensions—lexical, syntactic, semantic, and entity-level—making it difficult for any single feature type to capture the full signal. Even modestly performing features like named entity recognition (NER), which alone improves development accuracy from 0.894 to 0.900, play a critical role in disambiguation when combined with other signals, underscoring the value of multi-dimensional integration. 

This work makes several key contributions. First, we demonstrate that lightweight, interpretable models can achieve impressive performance through strategic feature fusion, challenging the notion that deep learning is necessary for advanced text classification. Second, we provide a systematic analysis of linguistic feature interactions, revealing why syntax and entity recognition offer complementary signals to distributional semantics. Third, we introduce an efficient architecture achieving a superior performance-interpretability-computation trade-off, particularly valuable for regulated industries, resource-constrained environments, and sustainability-minded applications. Finally, we establish a new benchmark for what transparent, feature-driven models can achieve when linguistic insights are thoughtfully applied.

LinguaSynth offers a principled and scalable alternative to the prevailing trend toward increasingly large and opaque neural models, demonstrating that deeper linguistic understanding—not just greater model size—can lead to more effective, interpretable, and resource-efficient NLP systems.

\section{Literature Review}

While deep learning has dominated recent NLP research, the field's pursuit of ever-larger models raises concerns about interpretability, computational efficiency, and environmental sustainability. This work explores an alternative path that prioritizes transparent feature engineering over black-box architectures. We review relevant literature across several dimensions that inform our approach.

\subsection{Limitations of Deep Neural Networks}

The NLP landscape has been transformed by transformer-based architectures, which achieve state-of-the-art performance across diverse tasks but at substantial costs. These models require extensive computational resources for both training and inference, with environmental impacts that have been quantified as equivalent to the lifetime emissions of multiple automobiles \cite{strubell2019}. More critically for many applications, their decision processes remain largely opaque, with hundreds of millions of unlabeled parameters making interpretation challenging \cite{lipton2018}.

These constraints—interpretability requirements in regulated domains \cite{rudin2019}, deployment limitations on edge devices, and sustainability concerns—have motivated research into more efficient, transparent alternatives. Our work contributes to this growing body of research seeking optimal trade-offs between performance and interpretability.

\subsection{Feature Representation Approaches}

\subsubsection{Lexical Features}

Term frequency-inverse document frequency (TF-IDF) remains a cornerstone of text representation due to its simplicity and interpretability. Each dimension in a TF-IDF vector corresponds to a specific word or n-gram, making the representation directly human-readable \cite{salton1988}. However, TF-IDF cannot capture semantic relationships between distinct but related terms, limiting its effectiveness on semantically nuanced tasks.

\subsubsection{Distributional Semantic Representations}

Word embedding approaches like GloVe \cite{pennington2014} address this limitation by learning dense vector representations through global co-occurrence statistics. These embeddings encode semantic relationships as vector arithmetic operations (e.g., king - man + woman $\approx$ queen) and have demonstrated significant improvements over count-based methods in various tasks. Document-level extensions such as Doc2Vec \cite{le2014} capture thematic structure by jointly learning document and word representations.

While distributional approaches effectively model semantic relationships, they sacrifice the direct interpretability of sparse lexical features. Moreover, as \citet{schnabel2015} demonstrated, embedding methods can perform worse than sparse representations on certain classification tasks where specific terminology is highly predictive.

\subsubsection{Syntactic and Entity-Level Features}
Part-of-speech (POS) features capture grammatical patterns that reflect stylistic and genre differences \cite{argamon2007}. POS histograms and n-grams have proven particularly effective for authorship attribution \cite{stamatatos2009} and genre detection \cite{feldman2009}, but are rarely integrated with distributional representations in a unified framework.

Named entity recognition (NER) features extract mentions of real-world entities (persons, organizations, locations), highlighting key actors and topics \cite{nadeau2007}. These features provide crucial disambiguation signals, distinguishing between references to companies versus products or identifying domain-specific actors.

\subsection{Hybrid Approaches and Feature Fusion}
Previous attempts at feature fusion typically explore limited combinations—often just two feature types—without systematic analysis of cross-feature interactions. \citet{yin2016} demonstrated improvements by combining multiple word embeddings but focused solely on semantic representations. Similarly, \citet{zhang2018} explored combining syntactic dependencies with distributional features but did not address entity-level information or full-spectrum linguistic fusion.

The few approaches that do incorporate diverse feature types often employ complex neural architectures that sacrifice interpretability. This creates a perceived trade-off between model complexity and interpretability that our work challenges.

\subsection{Research Motivation}
To our knowledge, no prior work has systematically combined all five complementary linguistic signal types—lexical (TF-IDF), syntactic (POS), entity-level (NER), word-level semantics (GloVe), and document-level semantics (Doc2Vec)—within a single, transparent architecture. Moreover, existing fusion-based models rarely examine the individual and joint contributions of each feature type or explain why specific linguistic signals offer complementary value.
\begin{enumerate}[itemsep=0pt, topsep=0pt, parsep=0pt, partopsep=0pt]
    \item Strategically integrating five linguistically diverse feature sets within an interpretable logistic regression framework
    \item Quantifying feature interactions and their contributions to classification decisions
    \item showing that thoughtfully engineered features paired with simple classifiers can achieve strong performance 
    \item Establishing a new benchmark in balancing performance and interpretability for text classification.
\end{enumerate}

By prioritizing feature complementarity over architectural complexity, LinguaSynth offers a compelling alternative for settings where model transparency, efficiency, and sustainability are critical—such as regulated industries, low-resource environments, and organizations mindful of computational cost.

\section{Methodology}
\subsection{Architectural Philosophy}
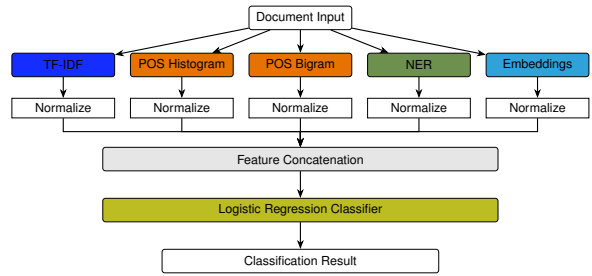
\begin{figure}[ht]
\centering
\resizebox{1\columnwidth}{!}{ 
\begin{tikzpicture}[
    box/.style={
      draw,
      rounded corners=2pt,
      minimum width=2.6cm,
      minimum height=0.6cm, 
      align=center,
      font=\sffamily\small
    },
    smallbox/.style={
      draw,
      minimum width=2.6cm,
      minimum height=0.5cm, 
      align=center,
      font=\sffamily\footnotesize
    },
    node distance=0.6cm]
    
    \node[box] (doc) at (0,0) {Document Input};

    \node[box, fill=tfidfcolor] (tfidf) at (-6,-1.2) {TF-IDF};
    \node[box, fill=poscolor] (poshist) at (-3,-1.2) {POS Histogram};
    \node[box, fill=poscolor] (posbigram) at (0,-1.2) {POS Bigram};
    \node[box, fill=nercolor] (ner) at (3,-1.2) {NER};
    \node[box, fill=embedcolor] (emb) at (6,-1.2) {Embeddings};

    \node[smallbox] (norm1) at (-6,-2.3) {Normalize};
    \node[smallbox] (norm2) at (-3,-2.3) {Normalize};
    \node[smallbox] (norm3) at (0,-2.3) {Normalize};
    \node[smallbox] (norm4) at (3,-2.3) {Normalize};
    \node[smallbox] (norm5) at (6,-2.3) {Normalize};

    \node[box, fill=lightrow, minimum width=10cm] (fusion) at (0,-3.6) {Feature Concatenation};
    \node[box, fill=hybridcolor, minimum width=10cm] (clf) at (0,-4.9) {Logistic Regression Classifier};
    \node[box, fill=white, minimum width=7cm] (result) at (0,-6.2) {Classification Result};

    \draw[arrow] (doc) -- (tfidf);
    \draw[arrow] (doc) -- (poshist);
    \draw[arrow] (doc) -- (posbigram);
    \draw[arrow] (doc) -- (ner);
    \draw[arrow] (doc) -- (emb);

    \draw[arrow] (tfidf) -- (norm1);
    \draw[arrow] (poshist) -- (norm2);
    \draw[arrow] (posbigram) -- (norm3);
    \draw[arrow] (ner) -- (norm4);
    \draw[arrow] (emb) -- (norm5);

    \draw[arrow] (norm1) -- ++(0,-0.6) -| (fusion);
    \draw[arrow] (norm2) -- ++(0,-0.6) -| (fusion);
    \draw[arrow] (norm3) -- (fusion);
    \draw[arrow] (norm4) -- ++(0,-0.6) -| (fusion);
    \draw[arrow] (norm5) -- ++(0,-0.6) -| (fusion);

    \draw[arrow] (fusion) -- (clf);
    \draw[arrow] (clf) -- (result);
\end{tikzpicture}
}
\caption{Overview of LinguaSynth.}
\label{fig:pipeline}
\end{figure}

LinguaSynth is grounded in a design philosophy that emphasizes feature complementarity and interpretability. Instead of relying on end-to-end representation learning, we explicitly engineer features that capture distinct linguistic dimensions and integrate them within a transparent, modular architecture. The overall pipeline is illustrated in Figure~\ref{fig:pipeline}.

This architecture offers several key benefits. It ensures full interpretability, as individual feature contributions can be directly analyzed via logistic regression weights. It is computationally efficient, with lightweight, parallelizable feature extraction requiring far fewer resources than deep learning models. The modular structure supports flexibility—features can be added, removed, or updated without retraining the full system. Additionally, robustness arises from the diverse feature set, which introduces redundancy and resilience against individual feature degradation.

\subsection{Linguistic Signal Extraction}
Table~\ref{tab:feature_architecture} outlines the six feature blocks used in the LinguaSynth architecture, along with their extraction methods, dimensionalities, and associated libraries. 
\begin{table}[ht]
\centering
\scriptsize
\begin{tabular}{@{\hskip 2pt}l@{\hskip 2pt}l@{\hskip 2pt}l@{\hskip 2pt}l@{\hskip 2pt}}
\toprule
\textbf{Feature Block} & \textbf{Extractor} & \textbf{Dimension} & \textbf{Library} \\
\midrule
\rowcolor{lightrow}
TF-IDF & TfidfVectorizer & $\approx$20{,}000 & sklearn \\
POS Histogram & POS tagger & 45 & NLTK \\
\rowcolor{lightrow}
POS Bigrams & POS tagger + TF-IDF & $\approx$1{,}500 & NLTK + sklearn \\
NER Histogram & Named Entity Recognition & 9 & spaCy \\
\rowcolor{lightrow}
GloVe & Pre-trained embeddings & 300 & Stanford NLP \\
Doc2Vec & Trained document vectors & 300 & Gensim \\
\bottomrule
\end{tabular}
\caption{LinguaSynth feature architecture details.}
\label{tab:feature_architecture}
\end{table}

\subsubsection{Surface Lexical Features: TF-IDF}
TF-IDF (Term Frequency-Inverse Document Frequency) captures word importance relative to document and corpus frequencies. After extensive optimization, we use 20,000 maximum features with unigram representation. This creates a sparse vector that effectively identifies distinctive terms while downweighting common words. 

For a term $t$ in document $d$ within corpus $D$, we calculate:

\begin{center}
$\text{TF-IDF}(t, d, D) = \text{TF}(t, d) \times \text{IDF}(t, D)$
\end{center}

Where TF is the normalized frequency of term $t$ in document $d$, and IDF downweights terms appearing in many documents.

\subsubsection{Syntactic Structure: Part-of-Speech Features} 
We capture document syntax using two complementary representations. The first, POS Histogram (global syntactic distribution), counts the frequency of 45 Penn Treebank tags per document, normalized to form a probability distribution. This 45-dimensional vector serves as a "grammatical fingerprint," capturing broad syntactic tendencies—such as noun versus verb prevalence—that help distinguish genres and writing styles.

The second, POS Bigram TF-IDF (local syntactic patterns), extracts consecutive POS tag pairs (e.g., "DT→NNP", "NNP→VBZ") and applies TF-IDF weighting to highlight distinctive local constructions. Dimensionality is reduced via truncated SVD, resulting in a 1,500-dimensional vector encoding fine-grained syntactic cues, such as noun–adjective patterns in technical texts or verb–noun sequences in sports narratives. 

Listing \ref{lst:pos_histogram} shows our implementation details of the POS histogram extractor.

{\tiny
\begin{lstlisting}[float=ht, caption={POS histogram implementation capturing a document's "grammatical fingerprint".}, label={lst:pos_histogram}]
SELECTED_TAGS = ["CC", "CD", "DT", "EX", "FW", "IN", "JJ", "JJR", "JJS", "LS", "MD", "NN", "NNS", "NNP", 
"NNPS", "PDT", "POS", "PRP", "PRP$", "RB", "RBR", "RBS", "RP", "SYM", "TO", "UH", "VB", "VBD", "VBG", "VBN", 
"VBP", "VBZ", "WDT", "WP", "WP$", "WRB", "#", "$", "``", "''", ",", ".", ":", "-LRB-", "-RRB-"]   

TAG2IDX = {t:i for i,t in enumerate(SELECTED_TAGS)}

def pos_histogram(doc: str):
    vec = np.zeros(len(SELECTED_TAGS), dtype=np.float32)
    for _, tag in nltk.pos_tag(nltk.word_tokenize(doc)):
        idx = TAG2IDX.get(tag)
        if idx is not None:
            vec[idx] += 1
    if vec.sum():
        vec /= vec.sum() 
    return vec

pos_hist_block = Pipeline([
    ("hist", FunctionTransformer(lambda docs: np.vstack([pos_histogram(d) for d in docs]), validate=False)),
    ("scal", StandardScaler()),
])
\end{lstlisting}
}

\begin{figure}[ht]
\centering
\centering
\includegraphics[width=0.6\linewidth]{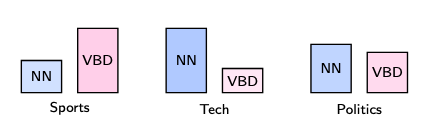}
\caption{POS patterns across domains. }
\label{fig:pos_distribution}
\end{figure}

\subsubsection{Entity-Level Understanding: Named Entity Features}
Named entity recognition captures the key actors, locations, organizations, and objects mentioned in text. For each document, we extract and normalize the frequency distribution of 9 entity types (PERSON, ORGANIZATION, LOCATION, etc.), creating a 9-dimensional "entity fingerprint" that encodes the types of real-world entities the document discusses. Listing \ref{lst:ner_histogram} provides the implementation. 
\begin{lstlisting}[float=ht, caption={Named entity histogram capturing a document's "entity fingerprint".}, label={lst:ner_histogram}]
NER_TYPES = ["PERSON","NORP","FAC","ORG","GPE","LOC", "PRODUCT","EVENT","DATE","TIME","CARDINAL","MONEY"]
NER2IDX = {lbl:i for i,lbl in enumerate(NER_TYPES)}

class NerHistogram(BaseEstimator, TransformerMixin):
    def fit(self, X, y=None): return self
    def transform(self, X):
        out = np.zeros((len(X), len(NER_TYPES)), dtype=np.float32)
        for i, doc in enumerate(_NLP.pipe(X, batch_size=32)):
            for ent in doc.ents:
                j = NER2IDX.get(ent.label_)
                if j is not None:
                    out[i, j] += 1
        row_sums = out.sum(axis=1)
        nonzero = row_sums > 0
        out[nonzero] = out[nonzero] / row_sums[nonzero, np.newaxis]
        return out

ner_block = Pipeline([
    ("hist", NerHistogram()),
    ("scal", StandardScaler())
])
\end{lstlisting}

\begin{figure}[ht]
\begin{minipage}[t]{0.48\textwidth}
\centering
\includegraphics[width=0.75\linewidth]{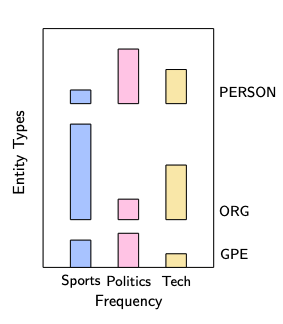}
\caption{Domain-specific entity trends.}
\label{fig:ner_distribution}
\end{minipage}
\hfill
\begin{minipage}[t]{0.48\textwidth}
\centering
\centering
\includegraphics[width=0.8\linewidth]{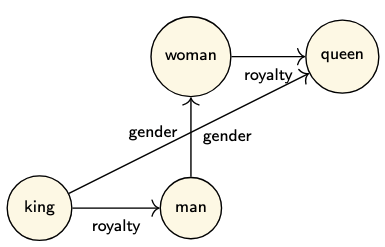}
\caption{GloVe vector space captures semantics via geometry.}
\label{fig:word_embeddings}
\end{minipage}
\end{figure}

\subsubsection{Distributional Semantics: Word and Document Embeddings}
We incorporate two complementary distributional representations. For word-level semantics, we use GloVe embeddings, which learn dense vector representations by factorizing a global word–co-occurrence matrix. This method places semantically related words (e.g., king – man + woman $\approx$ queen) near each other in vector space, as shown in Figure~\ref{fig:word_embeddings}. We use pre-trained 300-dimensional GloVe vectors and compute document-level representations by averaging word vectors, capturing semantic content independently of specific lexical choices. 

For document-level semantics, we use Doc2Vec embeddings, which extend word embeddings by learning fixed-length representations for entire documents. We train a 300-dimensional Doc2Vec model on our corpus, yielding embeddings that encode thematic structure, stylistic tendencies, and discourse-level features beyond the scope of word-level models.

\subsection{Strategic Feature Fusion}
The core insight behind LinguaSynth is that diverse linguistic signals provide complementary, non-redundant information. To ensure balanced contributions across feature types, we apply tailored normalization: dense features (POS histogram, NER histogram, GloVe, Doc2Vec) are z-score standardized, while sparse features (TF-IDF, POS bigram TF-IDF) are L2-normalized to prevent high-dimensional vectors from dominating.

After normalization, all feature blocks are concatenated into a single composite vector, jointly capturing lexical, syntactic, entity-level, and semantic information. This unified representation is passed to a logistic regression classifier with L2 regularization (C=1), ensuring interpretability and robust generalization.

The implementation of this fusion pipeline is shown in Listing~\ref{lst:feature_union}. 

\begin{lstlisting}[float=ht, caption={Feature fusion implementation demonstrating how multiple linguistic signals are combined.}, label={lst:feature_union}]
hybrid_model = Pipeline([
    ("features", FeatureUnion([
        ("tfidf", tfidf_block),
        ("glove", glove_block),
        ("d2v", doc2vec_block),
        ("pos_hist", pos_hist_block),
        ("pos_bi", pos_bi_block),
        ("ner", ner_block)
    ])),
    ("clf", LogisticRegression(max_iter=1000, C=1))
])

hybrid_model.fit(X_train, y_train)
\end{lstlisting}

\section{Experimental Design}

\subsection{Dataset and Preprocessing}
We conduct experiments on the 20 Newsgroups dataset, a standard benchmark containing approximately 18,000 newsgroup posts across 20 topic categories. Using the standard train/test split (11,314 training, 7,532 test documents), we further divide the training data into 80\% training and 20\% development for hyperparameter tuning.

\subsection{Evaluation Protocol}
We report macro-averaged accuracy as the primary evaluation metric to ensure equal weight across classes. For a more detailed analysis, we also examine per-class precision, recall, and F1 scores.

To isolate the contribution of each feature type, we conduct systematic ablation experiments, incrementally adding and combining components. All hyperparameters are tuned on the validation set and held constant across experiments to ensure fair and consistent comparisons.

\subsection{Experimental Configurations}
We evaluate a range of model configurations to assess the contribution of different feature types. First, we establish performance baselines using individual models—TF-IDF, GloVe, and Doc2Vec. We then explore syntax-enhanced models by adding POS histograms and/or bigrams to TF-IDF, and entity-aware models by incorporating NER features. Next, we evaluate a hybrid distributional model combining TF-IDF, GloVe, and Doc2Vec. Finally, we test the full LinguaSynth architecture, which integrates all feature blocks into a unified representation.

\section{Results and Analysis}

\subsection{Performance Comparison}

Table~\ref{tab:results} presents a comprehensive comparison of feature combinations across validation and test sets, revealing several key insights about how different linguistic signals contribute to classification performance. 

\begin{table}[ht]
\setlength{\tabcolsep}{3pt}
\centering
\scriptsize
\begin{tabular}{lrrcc}
\toprule
\rowcolor{lightrow}
\textbf{Model Configuration} & \textbf{Dev Acc.} & \textbf{Test Acc.} & \multicolumn{2}{c}{\textbf{vs TF-IDF Baseline}} \\
\cmidrule{4-5}
 & & & \textbf{Abs. $\Delta$} & \textbf{Rel. $\Delta$} \\
\midrule
TF-IDF + POS (Hist) & 0.8992 & 0.8139 & -0.0077 & -0.94\% \\
TF-IDF + POS (Bigram) & 0.9037 & 0.8174 & -0.0032 & -0.39\% \\
TF-IDF + POS (Both) & 0.9054 & 0.8186 & -0.0030 & -0.37\% \\
\rowcolor{lightrow}
TF-IDF + NER & 0.9001 & 0.8103 & -0.0113 & -1.38\% \\
TF-IDF + POS (Both) + NER & 0.9041 & 0.8200 & -0.0016 & -0.19\% \\
\midrule
\rowcolor{lightrow}
GloVe (Standalone) & 0.7698 & 0.7098 & -0.1118 & -13.61\% \\
Doc2Vec (Standalone) & 0.8471 & 0.7686 & -0.0530 & -6.45\% \\
\midrule
\rowcolor{highlightrow}
Hybrid (TF-IDF+GloVe+D2V) & 0.9213 & 0.8447 & +0.0236 & +2.93\% \\
\midrule
Hybrid + POS (Hist) & 0.9218 & 0.8441 & +0.0241 & +3.05\% \\
\rowcolor{highlightrow}
Hybrid + POS (Bigram) & 0.9196 & \textbf{0.8489} & \textbf{+0.0268} & \textbf{+3.26\%} \\
Hybrid + POS (Both) & 0.9191 & 0.8481 & +0.0265 & +3.23\% \\
\rowcolor{lightrow}
Hybrid + NER & 0.9183 & 0.8460 & +0.0244 & +2.97\% \\
Hybrid + POS (Both) + NER & 0.9178 & 0.8478 & +0.0262 & +3.19\% \\
\bottomrule
\end{tabular}
\caption{Feature combination comparison showing peak performance from linguistically diverse signal fusion.}
\label{tab:results}
\end{table}

To begin, TF-IDF provides a strong baseline at 82.16\% test accuracy, confirming that well-tuned lexical models remain highly effective. However, single-feature limitations quickly emerge. While POS and NER features slightly improve validation accuracy when added to TF-IDF, they reduce test performance, likely due to overfitting without sufficient complementary signals. For example, technical forums (e.g., comp.sys.ibm.pc.hardware, comp.windows.x) share similar noun-heavy POS profiles, limiting the discriminative value of POS histograms or bigrams in isolation.

In contrast, distributional semantics alone underperform, with GloVe (70.98\%) and Doc2Vec (76.86\%) individually lagging behind TF-IDF, indicating that pure semantic representations lose critical lexical specificity. This pattern is particularly pronounced for topics with tight, shared vocabulary like religious discussions (alt.atheism, soc.religion.christian), where TF-IDF alone captures most of the signal and distributional representations add minimal discriminative power. 

Among all single-feature augmentations, POS bigrams emerge as the most effective individual addition, boosting performance by +0.44 percentage points over the hybrid model and +3.73 over the TF-IDF baseline. These gains include precision improvements in 11 out of 20 categories and F1 score increases in 10, reflecting their ability to capture topic-specific syntactic patterns—such as noun sequences in hardware forums or verb–noun flows in sports content—that embeddings alone may overlook. In political texts, frequent transitions like Proper Noun → Verb (e.g., “Congress → debates”) and Determiner → Noun (“the → bill”) serve as stylistic markers that complement lexical features. 

Building on this, the hybrid distributional model excels, as the combination of TF-IDF, GloVe, and Doc2Vec reaches 84.48\% (+2.82\% over baseline), demonstrating that lexical and semantic features complement each other substantively. 

Finally, full LinguaSynth achieves peak performance. Adding POS bigrams to the hybrid model yields 84.89\% accuracy (+3.32\% over baseline), confirming the value of syntactic information even in the presence of rich semantic representations.

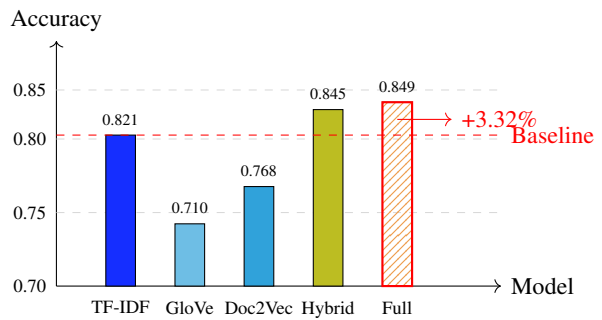
\begin{figure}[ht]
\centering
\begin{tikzpicture}[scale=0.65] 
    \draw[->] (0,0) -- (9,0) node[right, font=\small] {Model};
    \draw[->] (0,0) -- (0,5) node[above, font=\small] {Accuracy};
    
    \node[left, font=\scriptsize] at (0,0) {0.70};
    \node[left, font=\scriptsize] at (0,1.5) {0.75};
    \node[left, font=\scriptsize] at (0,3) {0.80};
    \node[left, font=\scriptsize] at (0,4) {0.85};
    
    \draw[gray!30, dashed] (0,1.5) -- (9,1.5);
    \draw[gray!30, dashed] (0,3) -- (9,3);
    \draw[gray!30, dashed] (0,4) -- (9,4);
    
    \draw[fill=tfidfcolor] (1,0) rectangle (1.6,3.08);
    \node[below, font=\scriptsize] at (1.3,-0.1) {TF-IDF};
    \node[above, font=\tiny] at (1.3,3.08) {0.821};
    
    \draw[fill=embedcolor!70] (2.4,0) rectangle (3,1.27);
    \node[below, font=\scriptsize] at (2.7,-0.1) {GloVe};
    \node[above, font=\tiny] at (2.7,1.27) {0.710};
    
    \draw[fill=embedcolor] (3.8,0) rectangle (4.4,2.03);
    \node[below, font=\scriptsize] at (4.1,-0.1) {Doc2Vec};
    \node[above, font=\tiny] at (4.1,2.03) {0.768};

    \draw[fill=hybridcolor] (5.2,0) rectangle (5.8,3.60);
    \node[below, font=\scriptsize] at (5.5,-0.1) {Hybrid};
    \node[above, font=\tiny] at (5.5,3.60) {0.845};
    
    \draw[fill=hybridcolor!70, pattern=north east lines, pattern color=poscolor!70] (6.6,0) rectangle (7.2,3.75);
    \node[below, font=\scriptsize] at (6.9,-0.1) {Full};
    \node[above, font=\tiny] at (6.9,3.75) {0.849};
    
    \draw[dashed, red] (0,3.08) -- (9,3.08) node[right, font=\small, red] {Baseline};
    \draw[->, red] (6.9,3.4) -- (8,3.4) node[right, font=\small, red] {+3.32\%};
    \draw[red, thick] (6.6,0) rectangle (7.2,3.75);
\end{tikzpicture}
\caption{Accuracy comparison of feature approaches, highlighting a 3.32\% improvement by the full LinguaSynth model over the TF-IDF baseline.}
\label{fig:results_vis}
\end{figure}

Figure~\ref{fig:results_vis} visually compares performance across feature combinations, highlighting LinguaSynth’s superior accuracy over individual approaches. 

\subsection{Feature Complementarity Analysis}

A key question is why these diverse linguistic signals work so effectively together. Our analysis identifies three core reasons for their complementarity, illustrated in Figure~\ref{fig:venn}.

First, each feature type captures an orthogonal linguistic dimension. TF-IDF captures \textit{what specific words} appear in the document, while POS features capture \textit{how language is structured} grammatically. NER features identify \textit{which real-world entities} are discussed, while GloVe captures \textit{what words mean} in a distributional sense. For example, in hardware forums, terms like “GPU” and “graphics card” are semantically close in GloVe space; in automotive discussions, “vehicle” and “car” cluster similarly—enabling the classifier to generalize beyond exact word forms. Doc2Vec adds a textit{overall document structure and theme} layer, capturing broader discourse structure and smoothing over varied phrasing to improve recall in cases where keyword overlap is low.

Second, we observe error diversity across feature types, enabling the model to recover from individual feature weaknesses. TF-IDF struggles with synonyms and lexical variation, while embeddings can conflate semantically similar words used in different domains. For instance, in the sentences “The game was exciting last night” and “This game has impressive graphics,” the word “game” refers to distinct contexts—sports versus video games. While GloVe embeddings may treat these as similar due to shared usage patterns, POS and NER provide orthogonal disambiguation signals—such as the presence of sports-related entities or technical modifiers—that help guide the classifier to the correct interpretation. Similarly, POS patterns may overlap across categories, and NER may miss or misclassify entities, but other features can compensate. This complementary error behavior is a key reason fusion models outperform single-feature approaches. 

\begin{figure}[ht]
\begin{minipage}[t]{0.47\textwidth}
\centering
\resizebox{\textwidth}{!}{%
\begin{tikzpicture}[scale=0.9]
  \draw[rounded corners, fill=white, draw=black] (0,0) rectangle (7,4);
  \node[align=center, font=\normalsize] at (3.5,3.5) {Example: The word ``game''};
  
  \draw[rounded corners, fill=tfidfcolor!20] (0.5,3.2) rectangle (6.4,1.8);
  \node[align=center, font=\small] at (3.5,2.45) {
    ``The game was exciting last night''\\
    $\downarrow$\\
    Sports context
  };
  
  \draw[rounded corners, fill=embedcolor!20] (0.5,1.6) rectangle (6.5,0.2);
  \node[align=center, font=\small] at (3.55,0.85) {
    ``This game has impressive graphics''\\
    $\downarrow$\\
    Computing context
  };
\end{tikzpicture}%
}
\caption{Contextual disambiguation of “game” aided by syntactic and entity-level signals.}
\label{fig:disambiguation_example}
\end{minipage}

\hfill
\begin{minipage}[t]{0.47\textwidth}
\centering
\resizebox{0.58\textwidth}{!}{%
\begin{tikzpicture}[scale=0.9]
  \draw[fill=tfidfcolor!40, fill opacity=0.5] (0,0) circle (1.3cm);
  \node at (0,0) {TF-IDF};
  
  \draw[fill=embedcolor!40, fill opacity=0.5] (1.9,0.9) circle (1.3cm);
  \node at (1.9,0.9) {Embeddings};
  
  \draw[fill=poscolor!40, fill opacity=0.5] (1.9,-0.9) circle (1.3cm);
  \node at (1.9,-0.9) {POS/NER};
  
  \node[font=\small] at (0.9,0.45) {lexical};
  \node[font=\small] at (0.9,-0.45) {structure};
  \node[font=\small] at (1.9,0) {context};
\end{tikzpicture}%
}
\caption{Linguistic feature complementarity across lexical, semantic, and structural dimensions.}
\label{fig:venn}
\end{minipage}
\end{figure}

Third, we observe disambiguation power. The combination of features often resolves ambiguities that single features cannot. For example, the word "game" appears in both sports and computer discussions, but sports articles show distinctive verb patterns (VBD-heavy) while computer discussions show different entity distributions (more PRODUCT mentions). When embeddings indicate semantic ambiguity, syntactic or entity patterns can provide disambiguating context.

This non-redundant signal is key to our approach—TF-IDF+GloVe+Doc2Vec capture what words mean, while POS/NER capture how they are used. Syntax highlights action-heavy versus description-heavy writing; entity counts hint at news-worthiness and domain-specific actors. Our analysis confirms that these lightweight linguistic features remain valuable even atop strong distributional embeddings, with POS-histogram still improving precision in 8/20 categories and recall in 5/20 categories, while NER-histogram features add modest but consistent lifts across the board (precision improvements in 8/20 categories).

\subsection{Category-Level Analysis}

To understand which categories benefit most from our approach, we examined per-class performance. Table \ref{tab:class_examples} shows examples of distinctive syntactic patterns found in different newsgroups.

\begin{table}[ht]
\setlength{\tabcolsep}{0pt}
\renewcommand{\arraystretch}{0.95}
\centering
\scriptsize
\resizebox{\columnwidth}{!}{
  \begin{tabular}{p{0.3\columnwidth}p{0.7\columnwidth}}
\toprule
\textbf{Topic} & \textbf{Distinctive Linguistic Patterns} \\
\midrule
\rowcolor{lightrow}
rec.sport.baseball & High VB, VBD (action verb), NNP (team/player name) \\
& \textit{"Martinez \textcolor{blue}{hit} the ball over the fence"} \\
& PERSON entities (players), high verb ratio \\
\midrule
sci.electronics & High NN, JJ (technical terms with adjectives) \\
& \textit{"The \textcolor{blue}{digital} \textcolor{red}{circuit} requires precise voltage"} \\
& PRODUCT entities, technical terminology \\
\midrule
\rowcolor{lightrow}
talk.politics.mideast & Complex NP, nested clauses (formal discourse) \\
& \textit{"The \textcolor{blue}{committee} on \textcolor{red}{foreign} \textcolor{green}{affairs} has decided..."} \\
& ORG and GPE entities, formal language patterns \\
\bottomrule
\end{tabular}
}
\caption{Category-specific linguistic signatures driven by syntax, vocabulary, and domain-specific entities.}
\label{tab:class_examples}
\end{table}

Our analysis reveals striking category-specific patterns that explain the effectiveness of multi-feature fusion. 

Sports articles are marked by high frequencies of action verbs (VBD), proper nouns (NNP), and PERSON entities, reflecting their focus on players, teams, and dynamic events. These texts exhibit a strong “action–noun” signature, which normalized POS distributions help highlight beyond surface-level word counts.

Technical categories show distinctive noun-adjective patterns and PRODUCT entities, reflecting detailed descriptions of technologies and specifications. Science and tech blend technical nouns with adjectives. The histogram flags that noun-adj balance, providing a syntactic signature that complements pure lexical features.

Political discussions feature complex syntax, including nested noun phrases, along with frequent mentions of GPE (geo-political entity) and ORG (organization). The global POS distribution provides a syntactic prior that boosts generalization, especially where lexical patterns vary across topics and phrasing. 

Religious texts contain more abstract nouns, fewer named entities, and a distinctive mix of determiners, pronouns, and theological terms. These shifts in syntactic structure and discourse markers are often missed by TF-IDF alone but captured effectively through POS-based features.

Overall, these findings highlight why feature fusion outperforms single-feature models: real-world categories exhibit distinctive signatures across lexical, syntactic, semantic, and entity-level dimensions that no individual feature type can fully capture.

\subsection{Interpretability-Performance Trade-off}
\begin{figure}[ht]
\centering
\begin{tikzpicture}[scale=0.8]
    \draw[->] (0,0) -- (3.5,0) node[right, font=\small] {Complexity};
    \draw[->] (0,0) -- (0,3.5) node[above, font=\small] {Performance};
    
    \node[circle, fill=tfidfcolor, draw=black, minimum size=0.3cm] (tfidf) at (0.8,1.7) {};
    \node[circle, fill=hybridcolor, draw=black, minimum size=0.3cm] (hybrid) at (1.7,2.5) {};
    \node[circle, fill=red!20, draw=black, minimum size=0.3cm] (dl) at (3,3) {};
    
    \draw[domain=0.6:3.2, smooth, variable=\x, black, dashed] 
      plot ({\x}, {0.3*\x*\x + 0.2*\x + 0.6});
    
    \node[above, font=\small] at (0.8,1.7) {TF-IDF};
    \node[above, font=\small] at (1.7,2.5) {LinguaSynth};
    \node[above, font=\small] at (3,3) {Deep Learning};
    
    \draw[rounded corners, red, dashed] (1.3,2) rectangle (2,2.8);
    \node[red, font=\footnotesize] at (1.65,1.7) {Our target};
\end{tikzpicture}
\caption{LinguaSynth's position in the performance–complexity space highlights its optimal trade-off for practical use.}
\label{fig:complexity}
\end{figure}
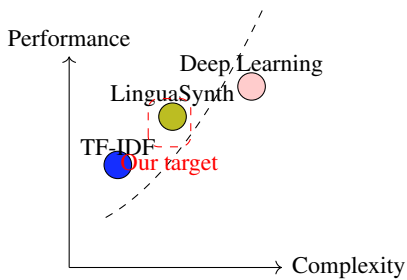

A crucial strength of LinguaSynth is its optimal balance between performance and interpretability. As shown in Figure~\ref{fig:complexity}, it occupies a strategic position between simple bag-of-words models and complex deep learning architectures.

LinguaSynth offers substantial advantages in this trade-off space. Training completes in minutes on standard hardware, and inference is performed in milliseconds—even on resource-constrained devices. The memory footprint is minimal, and feature extraction is highly efficient: POS and NER blocks (e.g., 45-dim POS histogram, 1.5k sparse POS bigrams, 9-dim NER) incur negligible computational cost compared to training or fine-tuning large language models.

Importantly, the model remains fully interpretable—logistic regression coefficients directly reveal each feature’s contribution to classification. Its modular architecture further supports maintainability, allowing for easy updates, domain adaptation, and debugging.

These characteristics make LinguaSynth especially well-suited for regulated industries (e.g., finance, healthcare, legal), resource-limited environments (e.g., edge devices, mobile applications), and sustainability-minded organizations seeking low-overhead, transparent AI solutions.

\section{Conclusion and Implications}
\subsection{Summary of Findings}
LinguaSynth achieves 84.89\% accuracy, outperforming a strong TF-IDF baseline by +3.32\% while maintaining full interpretability. By combining lexical features (TF-IDF), syntactic patterns (POS), named entities, and distributional semantics (GloVe, Doc2Vec), the model leverages complementary signals that no single feature type captures alone. Document categories exhibit distinct linguistic signatures across these dimensions, explaining the substantial performance gains from multi-source fusion. Importantly, syntactic and entity-level features contribute critical disambiguation, even when paired with strong semantic embeddings.

Almost every class benefits from our multi-faceted approach, underscoring that count-based (TF-IDF) and distributional semantic augmentation supply complementary signals that work together to improve classification performance.

\subsection{Broader Implications}
These results challenge the assumption that deep learning is necessary for high-performance text classification. We show that thoughtfully engineered linguistic features, combined with simple classifiers, can surpass strong baselines and deliver competitive accuracy. While end-to-end models dominate recent NLP research, our findings demonstrate that explicitly modeling linguistic structure offers both practical and theoretical advantages. LinguaSynth also contests the presumed trade-off between interpretability and performance, showing that they can coexist effectively.

\subsection{Practical Applications}

LinguaSynth offers a compelling alternative for numerous real-world scenarios. Regulated industries such as finance, healthcare, and legal benefit from our approach where model decisions must be transparent and explainable. Resource-constrained environments like mobile applications, edge devices, or low-resource languages gain efficiency without sacrificing accuracy. Applications requiring rapid deployment and iteration with limited computational resources can leverage our framework's lightweight nature, with training times measured in minutes rather than hours or days. 

Simplicity scales naturally in our framework, as all features use standard, off-the-shelf tools (NLTK, spaCy, Gensim) and fit comfortably on a laptop—ideal for low-resource or rapid-iteration settings. Organizations prioritizing environmental sustainability benefit from reduced carbon footprint compared to training large neural networks. Educational contexts gain value from understanding the linguistic basis of classification, enhancing learning and comprehension of NLP fundamentals.

Built-in interpretability is another core strength. Each feature is human-readable (e.g., verb ratio, PERSON count), making error analysis and domain adaptation straightforward. This transparency allows developers to trace misclassifications and make targeted improvements—an advantage rarely possible with black-box deep models.

\subsection{Future Directions} 
Several promising directions remain for future work. More sophisticated feature weighting strategies could optimize the relative contribution of each signal type beyond simple concatenation. Additional linguistic dimensions—such as discourse structure, pragmatics, or domain-specific features—may further enhance performance. Hybrid architectures combining interpretable models with lightweight neural components offer another path, potentially improving performance while preserving transparency. Finally, cross-domain and cross-lingual generalization studies would help assess the broader applicability and robustness of the LinguaSynth framework.

LinguaSynth sets a new benchmark for what interpretable, lightweight models can achieve. By strategically integrating complementary linguistic signals, we show that ever-larger neural networks are not the sole route to strong NLP performance. Instead, a deeper understanding of linguistic complementarity enables systems that are not only effective but also efficient and transparent—an essential consideration as AI continues to expand into sensitive and high-stakes domains.

\bibliographystyle{acl_natbib}
\bibliography{reference}

\end{document}